\documentclass[letterpaper, 10 pt, conference,table,xcdraw]{ieeeconf}  % Comment this line out if you need a4paper

\usepackage{amssymb} % per \checkmark
\usepackage{pifont}  % per \texttimes
\usepackage{booktabs}
\usepackage{comment}
\usepackage{url}
\usepackage{colortbl} % For \rowcolors command
\usepackage{graphicx}
\usepackage{hyperref}
\usepackage{amssymb} % Include the amssymb package for \checkmark symbol
\usepackage{tcolorbox}

\usepackage{adjustbox} % Include the adjustbox package
\usepackage[table,xcdraw]{} % For coloring tables

\IEEEoverridecommandlockouts                              % This command is only needed if 
\title{\LARGE \bf
Building Knowledge from Interactions: An LLM-Based Architecture for Adaptive Tutoring and Social Reasoning} %in HRI
\author{Luca Garello$^{*}$, Giulia Belgiovine$^{*}$, Gabriele Russo, Francesco Rea, Alessandra Sciutti%  <-this % stops a space
\thanks{*The first two authors contributed equally to this work.}% <-this % stops a space
\thanks{All authors are affiliated with the Italian Institute of Technology (IIT). Gabrielle Russo was affiliated with the University of Genoa at the time he contributed to this project.}
}

\begin{document}

\maketitle
\thispagestyle{empty}
\pagestyle{empty}

%%%%%%%%%%%%%%%%%%%%%%%%%%%%%%%%%%%%%%%%%%%%%%%%%%%%%%%%%%%%%%%%%%%%%%%%%%%%%%%%
\begin{abstract}
Integrating robotics into everyday scenarios like tutoring or physical training requires robots capable of adaptive, socially engaging, and goal-oriented interactions. 
While Large Language Models show promise in human-like communication, their standalone use is hindered by memory constraints and contextual incoherence.
This work presents a multimodal, cognitively inspired framework that enhances LLM-based autonomous decision-making in social and task-oriented Human-Robot Interaction. Specifically, we develop an LLM-based agent for a robot trainer, balancing social conversation with task guidance and goal-driven motivation.
To further enhance autonomy and personalization, we introduce a memory system for selecting, storing and retrieving experiences, facilitating generalized reasoning based on knowledge built across different interactions.
A preliminary HRI user study and offline experiments with a synthetic dataset validate our approach, demonstrating the system’s ability to manage complex interactions, autonomously drive training tasks, and build and retrieve contextual memories, advancing socially intelligent robotics.

\end{abstract}

%%%%%%%%%%%%%%%%%%%%%%%%%%%%%%%%%%%%%%%%%%%%%%%%%%%%%%%%%%%%%%%%%%%%%%%%%%%%%%%%
\section{INTRODUCTION}

%Current challenges and motivation
Developing robots that can assist and support humans in everyday tasks is a major challenge for the robotics community. Despite recent progress, robots are still not fully equipped to seamlessly coexist and interact with humans, due to the inherent unpredictability of human behavior.
% Why exploring LLM-based architectures
Large Language Model (LLM)-based architectures offer a promising solution for enhancing robotics and human-robot interaction (HRI) \cite{wang2024large, ZHANG2023100131}. Not only because they enable more natural, context-aware communication, but also for their potential to serve as a backbone for reasoning tasks, integrating different contextual information to perform abstract meta-reasoning and concepts summarization.
% GAPs in the HRI field
However, the literature still lacks a comprehensive exploration of LLM-powered architectures for embodied, interaction-focused applications \cite{incao2025roadmap}, especially in robots designed for goal-oriented tasks like education or tutoring, which demand both social engagement and adaptability.

LLMs struggle with reasoning about social norms, turn-taking, prolonged interactions, and delivering coherent, context-aware responses - key aspects of effective HRI \cite{irfan2025between}. Moreover, robots must autonomously manage interactions, efficiently store and represent the knowledge build through experience, and identify relevant information to personalize future encounters. Addressing these challenges requires enhancing LLM capabilities with advanced memory systems that emulate different types of memory (e.g., episodic, semantic, procedural), enabling robots to adapt, learn and respond more appropriately \cite{hatalis2023memory, zhang2024survey}.

To bridge these gaps, we propose an architecture that enhances a robot trainer’s social and decision-making abilities, enabling autonomous and adaptive behavior. % in physical training.
The tutoring context provides a valuable testbed for LLM reasoning, requiring both social engagement and task-oriented focus.  %regardless of the length of the interaction.
Our system enables the robot to autonomously guide interactions, choosing what to do next based on the context and balancing conversation and task completion to achieve training goals.
Building on this foundation, we extended our architecture to tackle knowledge representation challenges, enabling robots to store and retrieve interaction-based knowledge for future use. We opted for a Knowledge Graph (KG)-based approach \cite{pan2024unifying} for its explainability and ability to reveal relationships between entities, which other memory formats may overlook. Our long-term goal is to enable generalization and personalization across interactions, promoting adaptive long-term behavior.

In this work, we make the following contributions:
\begin{itemize}

    \item Development of an LLM-based agent for social and goal-oriented reasoning in a tutoring task.
     %\textbf{Research Question 1: Reasoning/Decision Making}: Can an LLM-powered architecture be used to enhance the reasoning and decision-making processes of robots to automate social and goal-oriented interactions in a tutoring context (e.g., deciding what should be done next, given a predefined goal and observed human behavior)?
    %The role of short-term memory (how to manage temporary, task-oriented information) to manage complex interactions to 1) develop autonomous robot's decision-making; 2) collect relevant information during such interactions (so that it can build  personalized behavior and interaction strategies later on)
        
    \item Design of a knowledge graph-based system, for representing robotic experiences and facilitating user-friendly exploration, and its validation for information retrieval and generation. %(how to best represent what the robot learned during a single interaction so that it can retrieve it easily in similar situations and generalize its knowledge in similar interactions with other users?)
% \textbf{Research Question 2: Memory and Knowledge}: Can an LLM-powered architecture enable robots to autonomously build general knowledge from their interaction with people and exploit this knowledge in future encounters?

    \item Integration of these components into a multimodal robotic architecture for autonomous HRI, with an initial evaluation via a user study.
    
\end{itemize}

\begin{figure}[!htbp]  % h = here, t = top, b = bottom, p = page, add optional placement
    \centering  % Center the image
    \includegraphics[width=0.47\textwidth, scale=0.4]{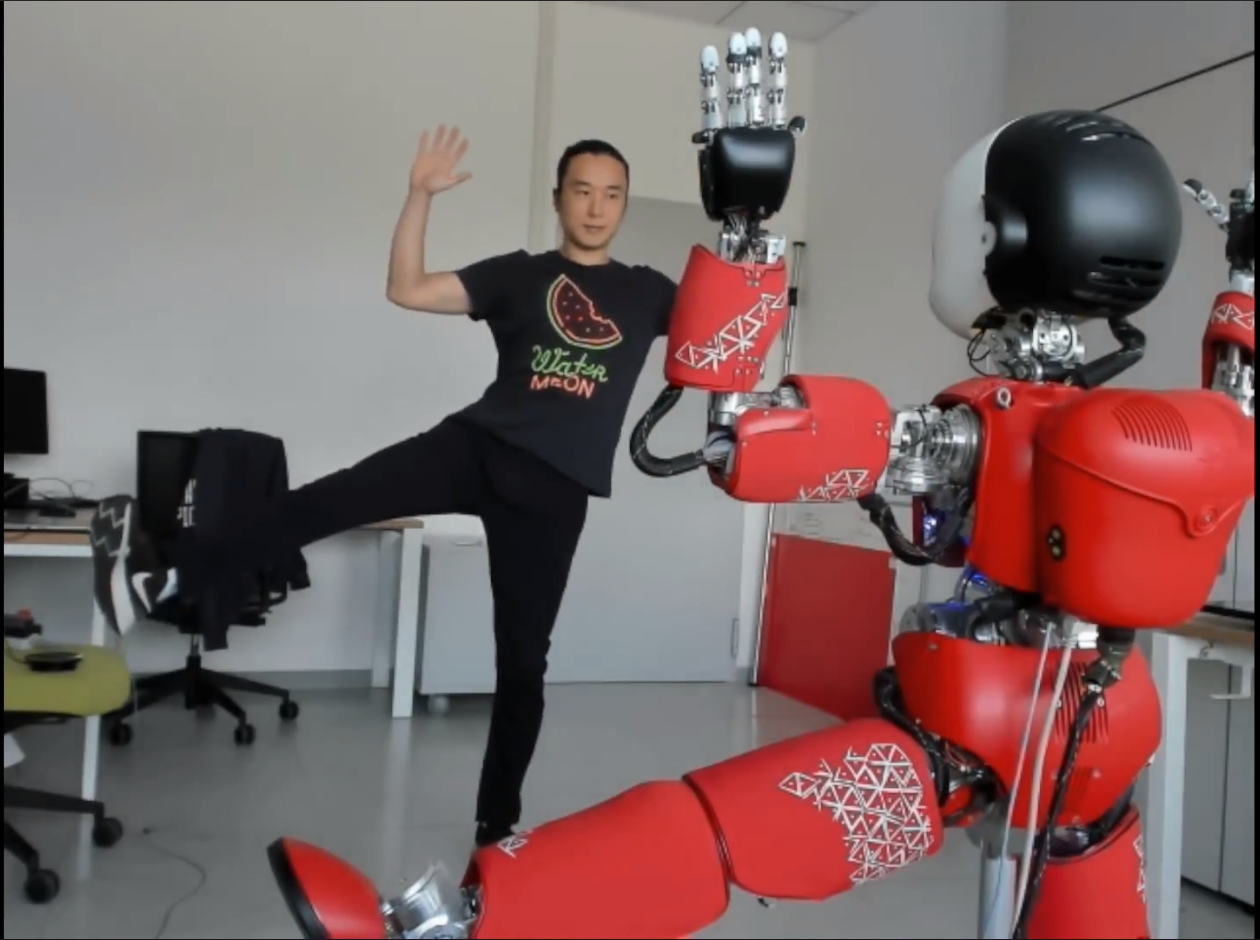}  
    \caption{HRI user study setup. The robot iCub showing a pose to a participant, providing real-time feedback.}  
    \label{fig:LLM_tools_list}  
\end{figure}

\section{RELATED WORK}
\label{relate_works:llm_hri}

% general advantage and limitation of LLMs in HRI
Integrating Large Language Models (LLMs) into HRI has greatly enhanced robots' ability to engage in natural conversations, handle diverse inputs, and support complex reasoning in interactive tasks.
However, LLM-powered robots introduce several challenges, including inconsistencies in logical reasoning, hallucinations, and increased user expectations for sophisticated non-verbal cues \cite{10.1145/3610977.3634966}. %, such as gestures, gaze coordination, and emotional expressiveness . 
Moreover, in the domain of Social Assistive Robots (SARs), an ideal system should continuously learn and reason about the user’s state to provide optimal assistance. The vast and continuos nature of user states and robot actions challenges traditional approaches like rule-based systems and reinforcement learning (RL) \cite{clabaugh2019escaping}, making the LLMs a promising alternative to foster strategic decision making by enabling a more complex representation of user states and actions.

%in the context of SARs and tutoring/coachig
Spitale et al. \cite{10.1145/3712265}, developed an LLM-powered mental health coach, combining informational and emotional goals for personalized interactions. Unlike our approach, the system uses the LLM solely for conversation,  with an RL module handling decision-making based on coaching state and user behavior. While it employs offline and adaptive RL for policy refinement, it lacks an explicit memory system, making it unclear whether the robot can recall past events or concepts.

In \cite{kang2024nadine}, the Nadine social robot utilizes an LLM agent (SoR-ReAct) for human-like cognitive and affective abilities. Like our approach, it processes multimodal data for user representation and employs tool-use capabilities. However, the tools serve as external resources (e.g., Google search) rather than domain-specific skills for autonomous interaction management. Another key difference is the memory system, which relies on retrieval-augmented generation (RAG) to dynamically access past interactions and knowledge at runtime.

Another approach that couple LLMs with a memory system is proposed by Ali et al. \cite{ali2024robots}. They introduce a hierarchical dual-layer memory system for an LLM-powered robotic architecture. It features working memory for real-time task tracking and declarative memory for storing past actions and environmental changes. A worker LLM manages memory retrieval, while a coordinator LLM integrates memory with inputs to guide task execution, enhancing adaptability, task switching, and context awareness.
Other studies \cite{maharana2024evaluatinglongtermconversationalmemory, hatalis2023memory} highlight the importance of structured memory types (e.g., short-term and long-term) to improve reasoning, retrieval accuracy, and response consistency of LLMs. %% ADD MORE CITATIONS IF YOU KNOW!!

\section{ARCHITECTURE}
%In this section, we present the \textit{Interaction Manager}, which serves as our LLM-based reasoning backbone, and the Memory System, a knowledge graph-based framework. We then describe how these modules are integrated into a comprehensive HRI architecture, seamlessly combining their functionalities with the robot’s additional perceptual and action capabilities.

\subsection{Interaction Manager}

The \textit{Interaction Manager} is an LLM-based agent designed to equip the iCub robot %\cite{metta2010icub} 
with reasoning-like capabilities, allowing it to autonomously guide users through a yoga training session without requiring human intervention or predefined plans. 

Its key functions include: (1) processing information derived from user conversation and context to determine the most appropriate next action; (2) managing interaction stage transitions while coordinating and communicating with relevant modules; and (3) tracking essential information to ensure coherent real-time behavior and trensferring relevant data to the long-term memory system. 

Unlike a conventional chatbot, the \textit{Interaction Manager} is not limited to passively responding to user inputs. Instead, it operates as a goal-driven entity, designed to guide the user through the training session while balancing social engagement with functional objectives, focusing on relevant episodes and concepts to build knowledge.

% Model and frameworks used
Due to the rapid evolution of LLMs during the study, we used GPT-3.5 Turbo for the HRI user study and GPT-4o Mini for offline evaluations, both accessed via the Azure OpenAI API. 
%We used LangChain{\href{https://python.langchain.com/docs/introduction/}{LangChain Documentation}} as the framework for structuring our LLM-based application.
To ensure safe system behavior, we implemented multiple safeguards, including built-in protections for sensitive contexts, adversarial testing during the pilot phase, and carefully crafted prompt engineering to regulate topic, tone, and appropriateness of conversations.

\subsubsection{Interaction Stages as Callable Tools}
\label{section:tools}

To develop a solution tailored to our robot tutor requirements, we built on the LangChain \footnote{\href{https://python.langchain.com/docs/introduction}{LangChain Documentation}} Agent concept, which rely on a supervising LLM with a system prompt instructing it to consult a set of customized tools before responding to user input (e.g., retrieving Wikipedia data). %While effective, this hierarchical setup introduces latency, making real-time interactions impractical for our use case.

\begin{figure}[htbp]  
    \centering  % Center the image
    \includegraphics[width=0.45\textwidth]{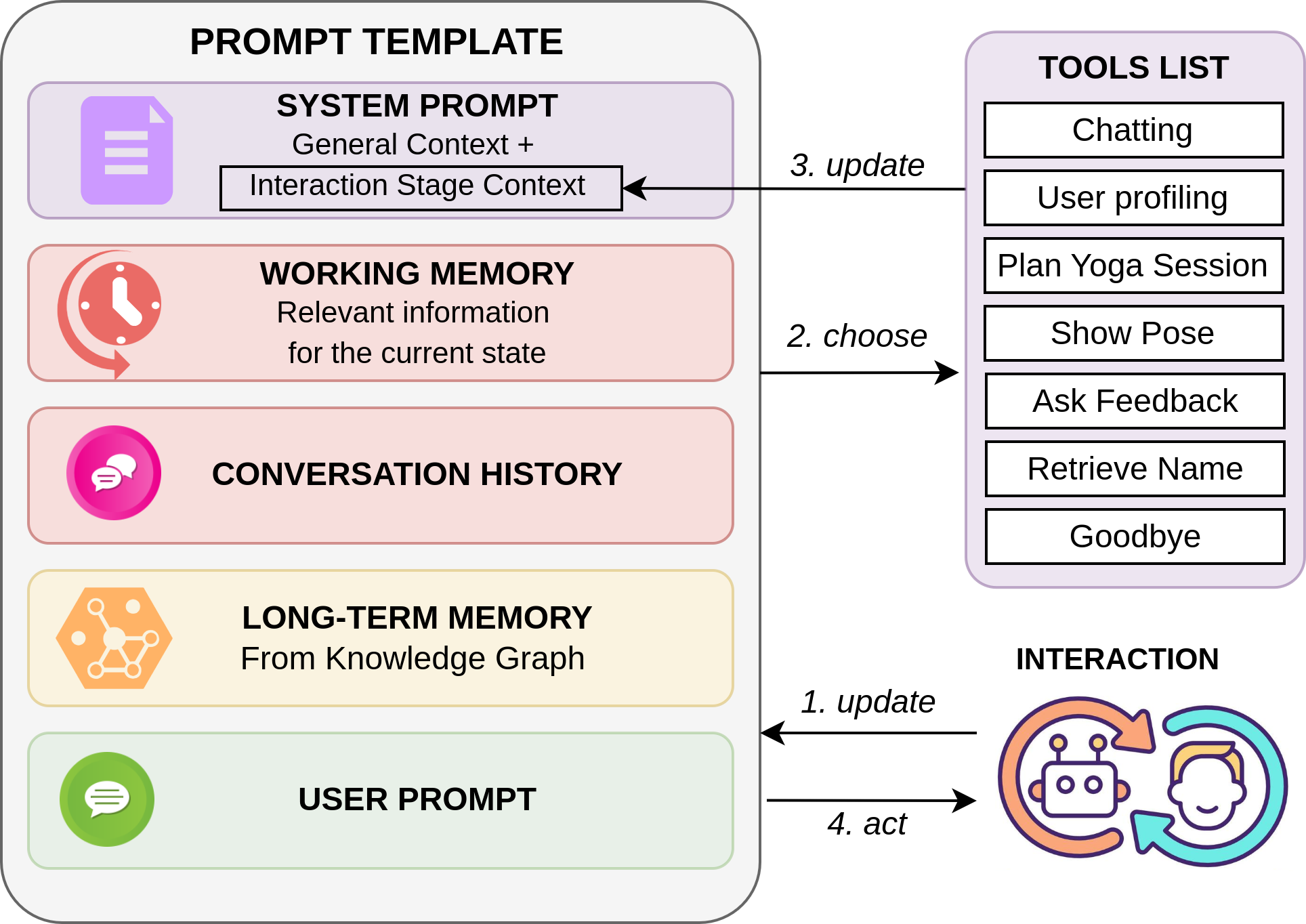} 
    \caption{LLM-Agent Prompt Template: during interaction, the robot collect information about the user's behavior (e.g., performance, prompts), updating the prompt template. The system determines whether to select a specific tool from the available list and updates the interaction stage and corresponding system prompt. Based on that, the robot executes specific verbal or motor behaviors.}  
    \label{fig:prompt}  
\end{figure}

%To address this, 
We optimized our approach by directly linking the LLM to custom tools, represented as OpenAI functions, each corresponding to a specific robot skill (e.g., user profiling, conversation, or training initiation), see Fig. \ref{fig:prompt}.
Using LangChain’s built-in functionality, we structured these tools within a prompt template, allowing the LLM agent to either respond directly or invoke tools as needed, triggering interaction stage transitions. This involves updating the system prompt with context-specific instructions and activating the necessary modules. This design ensures efficient real-time interactions, better suited to our requirements.

The toolset should be designed around domain-specific knowledge. In our case, we leveraged prior experience with this HRI use case \cite{belgiovine2022towards} to identify essential skills for an engaging and effective robot tutor. This design is iterative and evolves based on stakeholder needs and feedback.

%pipeline 
Once the LLM agent generates a response - consisting of an utterance, a tool invocation, or both - our pipeline updates the system prompt (if a new tool is activated) and the conversation history. To minimize latency and token costs, the conversation history is kept concise, retaining only the most recent 10 messages, a value chosen empirically. This approach ensures appropriate behavior, as the template prompt also includes a Working Memory component, which is further detailed in the next section (\ref{section:working_memory}). Finally, the verbal response is synthesized into audio using the Text-To-Speech module.

\subsubsection{Prompt Template and Working Memory}
\label{section:working_memory} 

%In addition to the current user prompt and the system prompt, our prompt template integrates a memory system to efficiently retain relevant information throughout the interaction.

As highlighted in previous studies \cite{maharana2024evaluatinglongtermconversationalmemory, hatalis2023memory}, relying solely on conversation history is often insufficient for effective agent reasoning and interaction, especially in long, multi-topic and multi-task exchanges \cite{ali2024robots}.
For this reason, we integrated a \textit{working memory} framework into the prompt, to efficiently retain relevant information throughout the interaction. 

In artificial agent architectures, \textit{working memory} refers to mechanisms that temporarily store information for ongoing tasks. Relevant data is eventually transferred to long-term memory, while short-term memory is cleared for the next interaction.
Additionally, this memory can function as a centralized repository, integrating multimodal data such as performance and conversational data, allowing the agent to make decisions based on a comprehensive real-time overview.

Our working memory framework addresses both temporal and centralized knowledge representation by capturing two types of information: (1) general details, such as the user's name, language, and training level, which must remain easily accessible; and (2) training-specific data, including pending and completed yoga poses, training status, and objectives. This information is essential for managing long conversations or interruptions during training, allowing the robot to seamlessly resume from where it left off when social moments temporarily shift the focus away from training.
%For instance, during the user profiling phase, the training status in working memory is set to \verb|training not started|. When the Reasoner module transitions from user profiling to the training phase, the status is automatically updated to \verb|training started|.

Finally, the structure of our template allows for the inclusion of information retrieved from the Knowledge Graph if specified (Fig. \ref{fig:prompt}).

%%%%%%%%%%%%%%%%%%%%%%%%%%%%%%%%%%%%%%%%%%%%%%%%%%%%%%%%%%%%%%%%%%%%%%%%%%%%%%%%%%%%%
\subsection{Knowledge-Graph based Memory System}

%\textbf{Final objective of this second solution: enabling the robot to autonomously build long-lasting knowledge from its experiences and interactions with different users.}
To enable personalized interactions with familiar users and analyze common patterns among successful and unsuccessful training sessions, a shared knowledge base (long-term memory) with efficient retrieval and update mechanisms is essential. While LLMs offer broad general knowledge, they lack the personal, context-specific insights robots gain through direct interaction.
To address this, we implemented a memory system that leverages LLMs to generate structured datasets of tables and text files (\textit{Memories dataset}), which are then transformed into a Knowledge Graph. This approach organizes the robot’s experiences into an interconnected network of knowledge and memories.
% Our hypothesis is that KG can be more efficient and explainable as a solution compared to other memory systems such as those based on RAGs.

\subsubsection{Interaction Memories Dataset}    

At the end of each interaction, our architecture autonomously build and updates a dataset of interaction memories. The dataset consists of the following files for each user:
\begin{itemize}
    \item \textit{Raw Data}: Includes the complete conversation between the robot and the user (``Raw\_Chat.txt"), along with raw performance data capturing joint values and errors for each frame processed by the \textit{Pose Extractor} system (``Raw\_Performance.csv").
    \item \textit{Meta Data:} Contains a conversation summary generated by the LLM, extracting key details such as users feedback on the training, hobbies, practiced sports, experience with yoga practice, etc. (``Summary.txt"). Additionally, it includes meta-performance data—a processed version of the raw performance—offering a more interpretable overview of the training session, such as average joint errors and the duration for which each pose was successfully maintained (``Meta\_Performance.csv"). These \textit{Meta Data} files serve as the foundation for building the KG and are designed to provide stakeholders, such as clinicians, with clear and actionable insights.
\end{itemize}
Finally, a log file is saved, capturing all technical details of the interaction. 

\subsubsection{Knowledge Graph} 
\label{section:methods_knowledge_graph}

At the end of each interaction, a user-specific KG is generated and linked to existing graphs through shared nodes (e.g., common sports or yoga poses among users). This process gradually builds a unified, evolving knowledge base as the robot gains experience.
The KG is built using LLMs (GPT-4o Mini)) and the Neo4j library, a graph database that employs the Cypher query language. Neo4j’s cloud-based service (Neo4j Aura) provides a graphical interface, enabling programmers and stakeholders to visually explore and query the robot’s knowledge.

To build the KG,
%we used the \textit{LLMGraphTransformer}, a LangChain transformer class to convert documents into graph structures using an LLM. 
we designed a system prompt with detailed instructions for the LLM to accurately extract nodes, relationships, and properties tailored to our use case. To minimize errors, the prompt includes predefined examples and a set of desired node and relationship labels (e.g., Person, Yoga Pose Performed (failed or completed), Sport, Hobbies, etc.). While this predefined structure ensures consistency, it remains flexible, allowing the LLM to generate new labels for user-specific concepts and unforeseen events.
For graph creation, we use \textit{Summary} files as the primary input, ensuring a more precise extraction of nodes and relationships while filtering out redundant or non-essential conversational data. Additionally, \textit{Meta\_Performance} data (e.g., performance scores, joints with the highest errors, etc.) is processed to enrich node properties further.

In HRI, robots are expected to be proactive, guiding interactions based on predefined goals and user needs. To support this, we designed a retrieval system that can be triggered not only by verbal queries but also by modalities like face recognition. When a user is identified, their identity is sent to the LLM agent, which generate a Cypher query for the graph. This retrieves the user node and its directly connected nodes, providing contextual information. However, at this stage, our implementation remains limited due to the absence of longitudinal user interactions.

Beyond personalization, an efficient retrieval system is key to identifying patterns across users. We evaluated these capabilities through offline studies, detailed in Section \ref{section:exp_memory}.

\begin{figure}[htbp]  % h = here, t = top, b = bottom, p = page, add optional placement
    \centering  % Center the image
    \includegraphics[width=0.35\textwidth, scale=0.10]{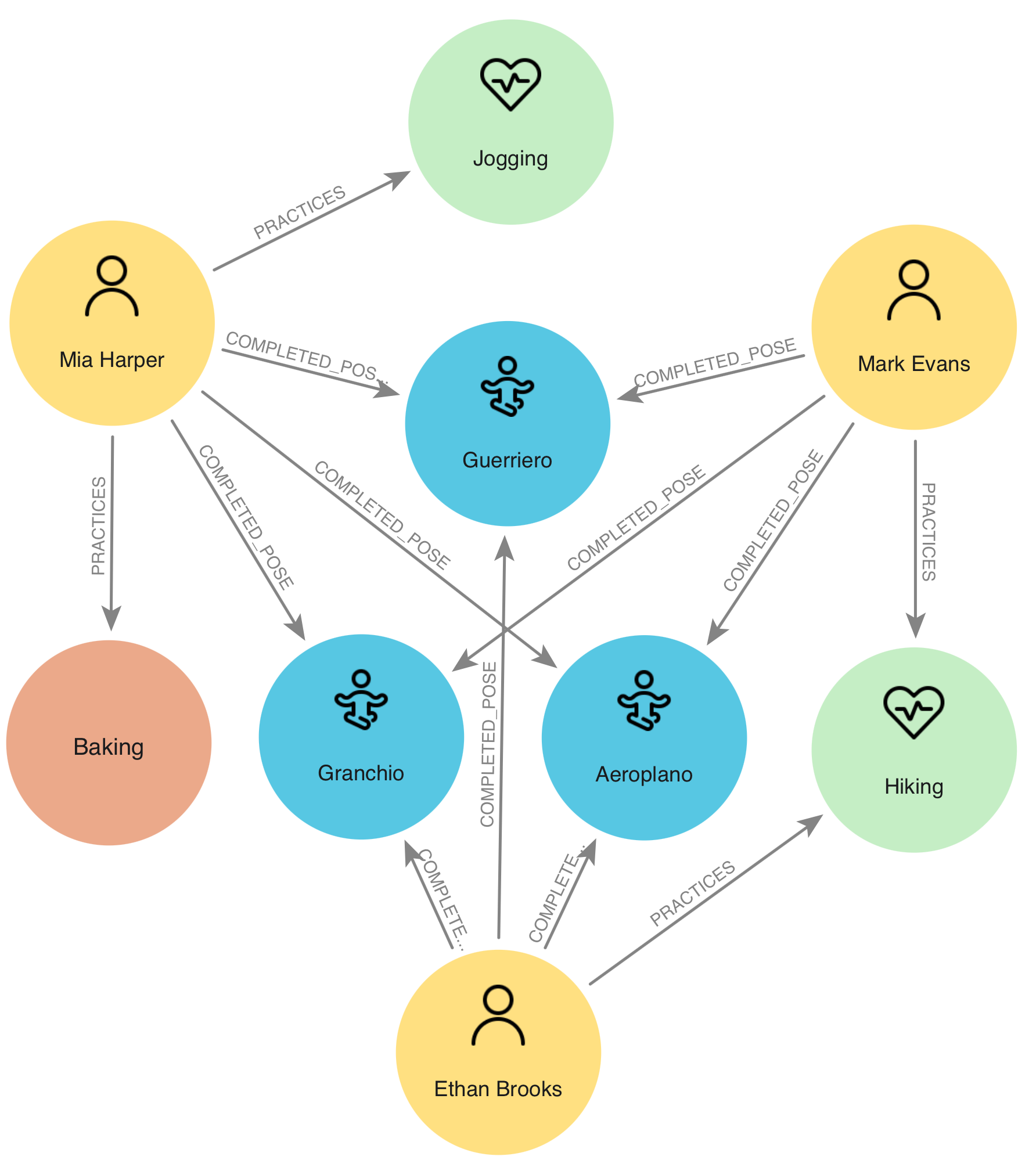}  
    \caption{Example of a section of the final graph, visualized through the Neo4j Aura app.}  
    \label{fig:graph}  
\end{figure}

%%%%%%%%%%%%%%%%%%%%%%%%%%%%%%%%%%%%%%%%%%%%%%%%%%%%%%%%%%%%%%%%%%%%%%%%%%%%%%%%%%%%%
\subsection{Integration and Implementation in HRI Architecture}

After designing the LLM agent for autonomous robot reasoning and developing the memory framework for knowledge representation and retrieval, we integrated them into a modular cognitive architecture tailored to our HRI tutoring scenarios.

%Artificial cognitive architectures model key human cognitive processes like perception, attention, memory, learning, and reasoning \cite{vernon2014artificial}. While developing a systems emulating human performance across all these capabilities remains a longstanding challenge \cite{kotseruba202040}, this work contributes to the field by proposing a framework for autonomous reasoning, decision-making, and interaction in personalized robotic tutoring. 

Our solution presents a modular design where each component is designed to perform a specific robotic function (Fig.~\ref{fig:Architecture}) . %, such as face detection, speech processing, or action planning. 
These modules communicate and exchange information via the YARP middleware \cite{metta2006yarp}. To improve resource allocation and optimize response times, the system employs a distributed approach. This design enables incremental updates, allowing new skills to be added by integrating additional modules into the framework over time. 

\subsubsection{Perception Modules}

These modules enable the robot to interpret its environment and interact with people by processing multimodal sensor data, by extracting high-level visual and audio features. Vision processing begins with RGB images at a resolution of 480×640 pixels, recorded at 30 fps by the robot’s cameras. The \textit{Face Recognition} relies on a face detector module, which utilizes the Ultralytics YOLOv8 model\footnote{\url{https://github.com/ultralytics/ultralytics}} to extract face bounding boxes. Face embeddings are subsequently computed and compared against a database to identify previously encountered individuals. 

The \textit{Pose Estimation} module generates a skeletal representation of the human body, consisting of 17 joints, using the YOLOv8-pose model from the Ultralytics library. For speech processing, the \textit{Speech-To-Text} module employs the state-of-the-art Whisper ``small-distill'' model\footnote{\url{https://github.com/openai/whisper}}, developed by OpenAI for highly accurate and robust transcription.

The pipeline detects the start of a new interaction either through user greetings or facial recognition. This information is then processed by the \textit{Interaction Manager} module, which determines whether to initiate the conversation with a greeting tailored to a known user or a more general introduction for an unknown user.

\subsubsection{Action Modules}

An \textit{Action Execution} module controls the robot’s physical movements, allowing it to demonstrate yoga poses and motor feedback as part of the training process. Some examples of yoga poses shown by the robot are described in \cite{belgiovine2022towards}. For neck and head movements, we employ the iKinGazeCtrl module - a controller designed for iCub's gaze, capable of independently steering the neck and eyes. Additionally, the action module provides functionalities for controlling facial expressions (by activating LEDs).
The \textit{Text-To-Speech} module converts text responses from the LLM into spoken messages, allowing the robot to communicate vocally. Robot voice is synthesized by using Acapela software\footnote{\url{https://www.acapela-group.com/}} with a child-like (Italian or English) voice.

\subsubsection{Trainer}

This module handles teaching-related tasks, including assessing user joint angles via the \textit{Pose Extractor} module.  It evaluates performance by comparing detected poses with expert-defined target poses, allowing for adjustable error tolerance (e.g., higher for children). When misalignment is detected, it provides corrective feedback through physical demonstration—repeating the correct movement—and verbal prompts suggesting posture adjustments. Further details can be found in \cite{belgiovine2022towards}.

\begin{figure}[!htbp]  % h = here, t = top, b = bottom, p = page, add optional placement
    \centering  % Center the image
    \includegraphics[width=0.5\textwidth, height=1.7in]{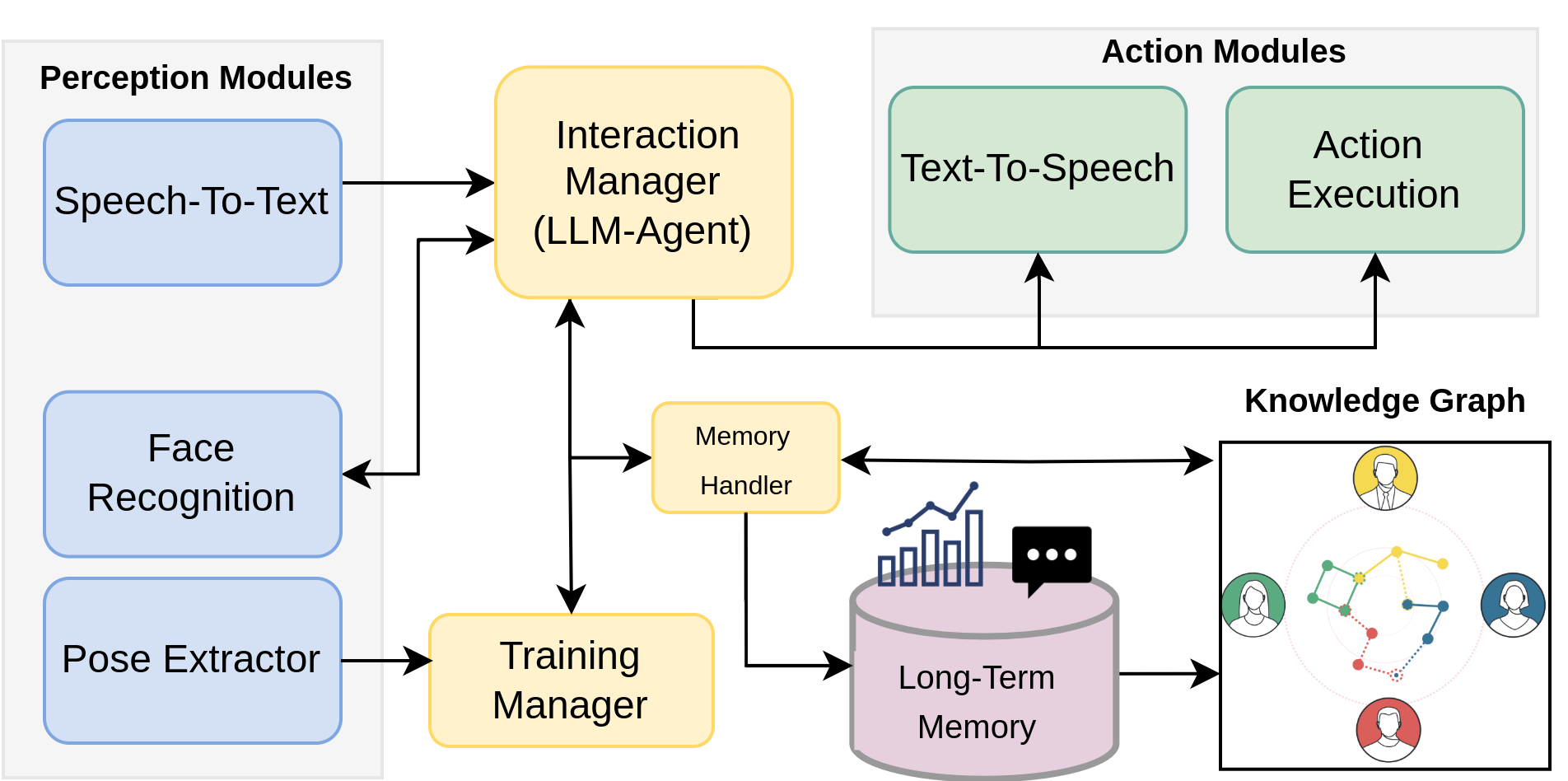}  
    \caption{HRI Architecture: perception modules process visual and audio data, which is sent to the LLM-based Interaction Manager for reasoning and conversation flow. The Trainer module handles performance assessment and feedback. Action modules control motor and verbal outputs, while Memory Handlers manage data retrieval and interface with the Knowledge Graph.}  
    \label{fig:Architecture}  
\end{figure}

%%%%%%%%%%%%%%%%%%%%%%%%%%%%%%%%%%%%%%%%%%%%%%%%%%%%%%%%%%%%%%%%%%%%%%%%%%%%%%%%%%%%%
\section{EXPERIMENTS}

\subsection{Architecture Evaluation through Real HRI User Study}
\label{section:exp_hri}

%\textbf{Is our architecture able to autonomously handle real-time human-robot interactions? How accurate is it?}

\subsubsection{User Study}
To evaluate our proposed architecture (Fig. \ref{fig:Architecture}), we conducted a user study to assess the system's effectiveness when fully integrated with the iCub humanoid robot for autonomous human-robot interactions.

\paragraph{Participant Demographics and Data Collection}
The study included 10 healthy volunteers, most of them unfamiliar with the robot. Participants ranged in age from 19 to 50 years (M=27 yo, SD= 8.8 yo) and were evenly split between 5 women and 5 men. As all were native Italian speakers, interactions were conducted in Italian. %Each interaction lasted an average of X minutes, and the entire data collection process spanned three days.

\paragraph{Experimental Protocol}
Participants interacted with the iCub robot, acting as a yoga instructor. To create a realistic testing environment, no specific restrictions or guidelines were imposed, aside from instructing participants to interact with the robot naturally.
Instead of formal questionnaires, we conducted semi-structured interview after each session to gather qualitative feedback on the robot’s social and tutoring behavior. The insights obtained were essential in identifying limitations and evaluating the system's overall performance.

\subsubsection{Evaluation Metrics}
We defined key metrics to evaluate the architecture's performance in autonomously allowing a successful yoga session with the robot. These metrics assessed the effectiveness of the integrated system, focusing on the LLM agent’s functionality and coordination with other modules.

We report three key metrics:  the \textit{Average Robot Response Time}, the \textit{Interaction Success Rate}, and the \textit{Architecture Success Rate}. %Additionally, we report the \textit{qualitative feedback} from participants regarding the robot's responses and the balance between its social and functional (tutoring-related) behavior.

The \textit{Average Robot Response Time} measures the robot's responsiveness during interactions, crucial for ensuring a natural and seamless user experience.

The \textit{Interaction Success Rate} evaluates the reasoning capability of the developed LLM agent in managing the interaction pipeline. It assesses the agent's ability to select the most appropriate tool and determine the correct stage transition based on context.

The \textit{Architecture Success Rate} evaluates the proper functioning and communication of all modules within the architecture (e.g., visual processing or speech-related functions). While these modules are not the study’s primary focus, their failures are reported to provide a holistic view of interaction quality and identify areas for improvement.
%Finally, we interviewed participants at the end of each interaction to gain their feedback about quality of responses and robot behaviour (e.g., the balance between its social and functional behavior). 
We report the primary reasons for architecture failures and users' feedback in Section \ref{results:architecture}.

%Important to report: Time and Hardware Costs (for retrieval, processing, and inference)

\subsection{Memory System Evaluation through Virtual HRI User Study} 
\label{section:exp_memory}

\paragraph{Fake Users Synthetic Dataset}
For a more comprehensive evaluation of the memory creation and retrieval pipeline, we generated a synthetic dataset of 28 users, hereafter referred to as fake users.

Each fake user is characterized by a bio detailing their name, age, interests, sports activities, profession, and personality profile. %An example of a fake user bio is provided in \ref{appendix:bio_fake_users}. 
The bios were generated using GPT-4o Mini with a carefully designed prompt including a predefined template and an example created by the authors. The prompt specified that the generated bios should follow a predefined age distribution and introduce variability in sport practices and levels. Personality traits, inspired by the big 5 personality traits model \cite{mccrae1992introduction}, were incorporated to ensure diversity in conversation styles.

Each user’s bio was provided as input to a second LLM agent, prompted to impersonate the described profile and to interact with the yoga tutor LLM-agent.                                                               
To simulate user performance (nodes attributes in our KG), each fake user’s bio included a success probability for yoga poses based on their sports habits. The boolean variable ``pose\_complete'' was determined by a weighted probability based on that information, while the most critical joint was selected randomly.

\paragraph{Summaries and Knowledge Graph Evaluation}

The first aspect we evaluated was the quality of the generated summary files, as they serve as the foundation for constructing the KG. This assessment was conducted empirically by the authors, who tested different prompt structures and analyzed the summaries in terms of clarity, fluency, and inclusion of relevant and correct information. We then manually examined the graphs built from these summaries, assessing the completeness and correctness of the information represented in the nodes and edges.

Next, we evaluated the quality of graph-based responses in retrieval task, using two distinct question types: (1) user-specific queries and (2) general queries requiring reasoning over the entire population. The first category included five standardized questions per user, covering their practiced sport and frequency, interests, training level, successfully completed poses, and feedback on the training. The second category involved queries that required reasoning over relationships between nodes, ranging from general questions requiring node counting (e.g., \textit{“How many people have successfully completed three poses?”} or \textit{“How many people currently practice swimming?”}), while others required more complex multi-step reasoning, such as \textit{“Which poses are most likely to be failed by beginners?”} or \textit{“Are users interested in X also interested in Y?”}.

To quantitatively assess response quality, we employed as evaluation metric the \textit{Faithfulness} score from the Ragas library\footnote{\url{https://docs.ragas.io/en/stable/}} which measures whether the response is grounded in the retrieved context (range: 0-1, 1 is ideal). The metric require a ground truth as a reference for evaluation. For user-specific questions, we used the corresponding user’s summary as the ground truth. For population-wide questions, a dedicated ground truth was manually curated for each question by the authors. Due to space limitations, the full list of questions is not included here but is available in the GitHub repository \footnote{\url{https://gitlab.iit.it/cognitiveInteraction/yoga-teacher-llm}}.
% \begin{itemize}
%     \item \textit{Faithfulness}: Measures whether the response is grounded in the retrieved context (range: 0-1, 1 is ideal).
%     %\item \textit{Factual correctness}: Assesses whether the response is factually accurate compared to the reference (range: 0-1, 1 is ideal).
%     % \item \textit{Response relevancy}: Evaluates whether the response directly addresses the user’s question (range: 0-1, 1 is ideal).
%     % \item \textit{Noise sensitivity}: Tests the robustness of the response against irrelevant or misleading context (range: 0-1, 0 is ideal).
% \end{itemize}

As retrieval method we used 2 approach the we call \textit{Naive Graph} and \textit{GraphCypherQA} method. In the \textit{Naive Graph} approach, we used a Cypher query to extract the subgraph corresponding to the individual’s data for user-specific questions, and a Cypher query to extract the whole graph for population-wide questions. In the \textit{GraphCypherQA} method we used an agentic LLM to directly translate the verbal query into a Cypher query, retrieving only the relevant information from the graph. 

%For user-specific queries, we employed a Cypher query to extract the subgraph corresponding to the individual’s data. For population-wide queries, the entire knowledge graph was retrieved. The resulting graph data were subsequently transformed into a text-based representation suitable for natural language processing. This text representation was then processed using the GPT-4o Mini model to generate the output response. To enhance the reliability and consistency of the model’s outputs, the prompt explicitly instructed the model to adhere closely to the phrasing and structure of the input question, minimizing extraneous information unless such elaboration was explicitly requested.

\paragraph{Comparison with RAG baseline}

We compared our graph-based approach with a baseline using a Retrieval-Augmented Generation (RAG) system. A RAG system typically combines a vector database with a LLM \cite{gao2023retrieval}. The vector database stores and retrieves contextual information relevant to user queries, while the LLM generates responses based on the retrieved context. While this method is effective in many scenarios, it encounters challenges with complex tasks such as multi-hop reasoning or answering questions that require integrating multiple pieces of information. 
Moreover, the black-box nature of vector representations and vector search makes it impossible to explain the origins of the gathered information. Consequently, users, stakeholders, and developers have limited insight into why specific chunks were retrieved and how they influence the generated response. This lack of explainability can be a major disadvantage, especially in fields such as healthcare, where transparency and accountability are essential.

To implement the RAG baseline, we developed a vector store that indexes user summaries conversations (the same used for building the graph). The conversation data is first segmented into smaller, overlapping chunks (chunk size = 500, chunk overlap = 50) to improve retrieval accuracy. %using a RecursiveCharacterTextSplitter 
Each chunk is stored as a document instance containing metadata, including the conversation ID and the associated user name. The vector representations of these documents are generated using the pre-trained text-embedding-ada-002 model by Azure OpenAI and stored in a FAISS vector database, which is saved locally for future use.
Similarity search is applied to retrieve most relevant context.

To evaluate the performance of the RAG system, we applied the same metric used for assessing the KG answers, as described in the previous section. Moreover, since the KG may include additional information derived from Meta Performance data (as explained in \ref{section:methods_knowledge_graph}), we ensured a fair comparison by selecting evaluation questions that could be answered using the information available in both the RAG and Knowledge Graph systems.

%Our hiphotesys was that the rag will not be good as tha KG in answering more general questions, that captures the relations and connections between different entities

%%%%%%%%%%%%%%%%%%%%%%%%%%%%%%%%%%%%%%%%%%%%%%%%
\section{RESULTS}

\subsection{Architecture Evaluation through Real HRI User Study}
\label{results:architecture}
% \textit{Is our architecture able to autonomously handle real-time human-robot interactions? How accurate is it?}

\textit{Interaction Success Rate} is 100\%, meaning that our LLM Agent always choose the right interaction stage (e.g., tool) to switch to.
The \textit{Average Robot Response Time} is 1.1 $\pm$ 0.26 s.
\textit{Architecture Success Rate} is 60\%, with 4 failures out of a total of 10 interactions.  %explain in detail which kind of failures
The architectural failures were caused by the following issues: errors in the speech synthesizer module when processing excessively long sentences (2),  failure to save the summary file due to an OpenAI API rate limit error (1), failure to save the person's facial data in the database due to low-quality vision input. (1).  These failures represented minor inconveniences in the operation of auxiliary modules which did not hinder the overall progress of the interaction. By analyzing these metrics, we confirmed the accuracy of our LLM agent and identified opportunities to enhance the performance of other modules.
Summary of the metrics and interaction failures can be found in Table \ref{tab:ExperimentResultsTable}.

Participants generally provided positive feedback, finding the yoga session engaging and appropriately challenging. However, natural interaction with the robot was hindered by response times that, while in line with average artificial agents response, fell short of human-like fluidity. Many users grew impatient, repeating phrases under the assumption the robot hadn’t heard, which overloaded the response system. Improvements in the robot’s non-verbal cues are needed to clearly convey that it is listening and processing inputs.
The quality and variety of responses were well-received. However, opinions varied regarding the balance between a ``chatty'' robot and a ``functional'' one focused on training objectives. Introverted or task-oriented participants appreciated the robot's concise responses, while extroverted or curious users found its tendency to avoid deep personal engagement - and redirect focus to training - somewhat rude. This dynamic warrants further investigation in future studies and highlights the potential for tailoring the robot’s behavior based on user interaction preference and personality.

%Another critical area for improvement is the agent's ability to handle jokes and sarcasm, particularly during training sessions.

\begin{table}[!h]
    \centering
    \caption{Experiment Results Summary}
    \label{tab:ExperimentResultsTable}
    
    \renewcommand{\arraystretch}{1.2}
    \setlength{\tabcolsep}{3pt} % riduce spazio tra colonne
    \rowcolors{3}{gray!15}{white}
    \footnotesize % font più piccolo per stare nella colonna IEEE
    \begin{adjustbox}{width=0.5\textwidth} % mezza pagina
    \begin{tabular}{c c c c c} % l'ultima colonna ha larghezza fissa con wrap automatico
        \toprule
        \textbf{\#} & \textbf{Avg.} & \textbf{Interaction} & \textbf{Architecture} & \textbf{Reason} \\
        \textbf{Subject} & \textbf{Resp. Time (s)} & \textbf{Success} & \textbf{Success} & \textbf{of Failure}\\
        \midrule
        
        1  & 0.75 & \checkmark & \checkmark  & - \\
        2  & 1.00 & \checkmark & \checkmark  & - \\
        3  & 1.15 & \checkmark & \texttimes & OPENAI rate limit\\
        4  & 1.18 & \checkmark & \checkmark  & - \\
        5  & 1.13 & \checkmark & \texttimes & TextToSpeech Failure\\
        6  & 1.13 & \checkmark & \texttimes & TextToSpeech Failure\\
        7  & 0.84 & \checkmark & \texttimes & Face not saved \\
        8  & 1.77 & \checkmark & \checkmark  & - \\
        9  & 1.07 & \checkmark & \checkmark  & - \\
        10 & 1.14 & \checkmark & \checkmark  & - \\
        \bottomrule
    \end{tabular}
    \end{adjustbox}
\end{table}

\subsection{Evaluation of Memory System through Virtual HRI User Study}

\paragraph{Quality Check of Summaries and Knowledge Graph}

After initially experimenting with a prompt that generated a more narrative summary format, we ultimately adopted a structured approach, where key concepts are categorized into distinct sections. 

The sections included in the structured summaries are: 1) \textit{General information}, collected by the Working Memory, such as the user's name, language, and the poses they successfully completed or failed; 2) \textit{Insights about user interests and activities}, extracted from the raw conversation and covering aspects like the user’s hobbies, current sport activities, past sport experience, and any prior or ongoing engagement with yoga; 3) \textit{Key aspects related to the user's training} (from raw chat as well), including specific challenges encountered during training, feedback provided at the end of the session, and any other relevant details that may contribute to a richer understanding of the user’s experience.
This method proved to be more effective in reducing ambiguity and managing complex or overlapping topics, such as discussions about yoga experiences outside the training session. 

As a result, the generated Knowledge Graph accurately captures all the required information while maintaining the intended structure in node and edge properties, ensuring factual correctness.

\subsection{Comparison with RAG baseline}

\begin{figure}[htbp]  % h = here, t = top, b = bottom, p = page, add optional placement
    \centering  % Center the image
    \includegraphics[width=0.51\textwidth]{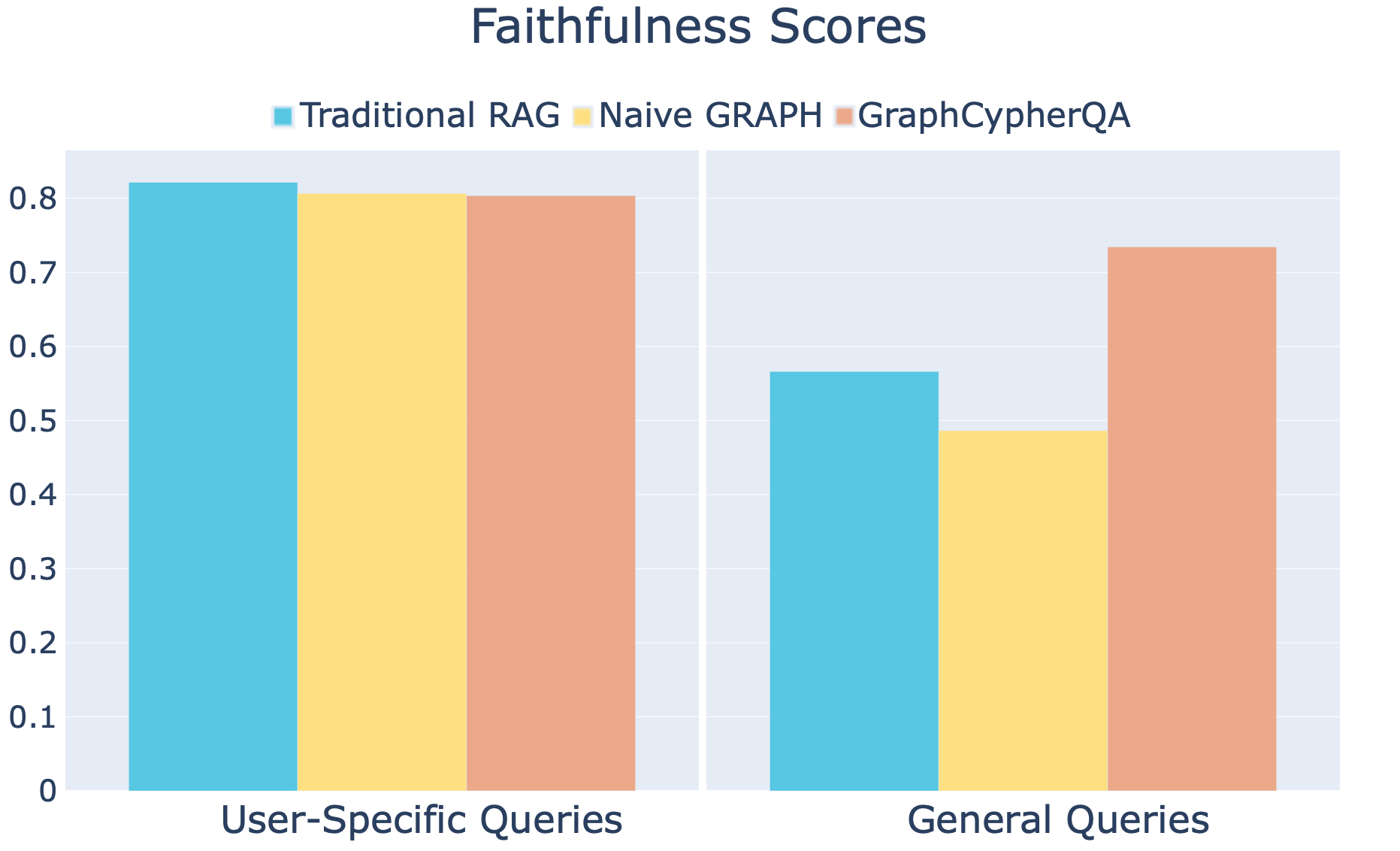}  
    \caption {Comparison of different retrieval methods for answering user questions. For specific questions about individual users, the performance of all methods is comparable. However, for complex, multi-hop questions requiring extended reasoning across the entire dataset — such as 'Who is the best practitioner in the database?' — the Cypher based approach outperforms other methods, demonstrating superior ability to aggregate and infer information across multiple data points."}  
    \label{fig:barplot-general}  
\end{figure}

For user-specific responses, the \textit{Faithfulness} scores were as follows: RAG achieved 0.82, Naive Graph scored 0.81, and GraphCypherQA reached 0.80. For general queries, the \textit{Faithfulness} scores were 0.57 for RAG, 0.49 for Naive Graph, and 0.73 for GraphCypherQA. See Fig. \ref{fig:barplot-general}.
While we did not observe a significant difference among the three methods for user-specific questions, the GraphCypherQA method outperformed the others in handling general, population-wide queries. This confirms that in cases requiring multi-hop reasoning, the graph-based approach offers distinct advantages, as widely supported by existing literature. The relatively small size of our dataset may have contributed to the differences being less pronounced.

\section{DISCUSSION}

Our study present the development of an LLM-based reasoning agent with a Knowledge Graph-based memory system and their integration in a multimodal, cognitively inspired framework to enhance autonomous decision-making in Human-Robot Interaction (HRI). Specifically, we developed a robot tutor capable of balancing social interaction with task-driven guidance in physical training scenarios. The results of our user study and offline evaluations highlight the system’s strengths, limitations, and potential for further improvement.

% Key findings from HRI user study (technical aspects)
The findings from the user study demonstrate that our architecture successfully enables real-time, autonomous decision-making during training interactions. The LLM-based agent consistently selected the appropriate interaction stage (or tool), achieving a 100\% \textit{Interaction Success Rate} with an average response time of 1.1 seconds.
While this latency is reasonable, users unfamiliar with vocal interactions with artificial systems sometimes misinterpreted pauses as comprehension failures, leading to repeated inputs and system overload. This issue underscores the need in achieving a fluent conversational flow by pushing on more intuitive non-verbal behavior, as suggesed also by \cite{10.1145/3610977.3634966}. Future iterations should incorporate cues like gaze shifts, gestures, or active listening indicators to improve interaction fluidity.

%Domain-Specific Tool Selection: 
In the proposed solution, the architecture’s reasoning capabilities rely on the availability of well-defined tools. The current toolset was designed based on prior HRI research in physical training, but adapting the approach to other domains, such as education or collaboration, may require further refinement. This process should involve collaboration among developers, users, and stakeholders to ensure tool and prompt design remains contextually relevant. Their active participation is essential for defining the appropriate level of robot autonomy and refining its action space'
%, potentially integrating explicit human intervention while maintaining reliable autonomous execution.

% auxiliar modules
Analyzing the architecture’s performance revealed areas for improvement, for auxiliary speech and vision modules. While these issues did not directly hinder task completion, they underscored the need for greater system robustness. Since human-robot interactions often occur in noisy, unstructured environments, future work should focus on developing resilient strategies for handling poor-quality input, such as implementing proactive behaviors like prompting users to move closer or requesting sentence repetition when faced with uncertainty.

% Interaction design aspects
Interviews after user study highlighted key insights into human preferences when engaging with social robots. While the conversational capabilities were well-received in terms of variety, fluidity, and response grounding, there were mixed opinions about the ideal trade-off between a chatty, engaging robot and a functional, task-focused one. Preferences largely depended on individual personality traits, suggesting that future iterations should incorporate adaptive interaction styles to enhance user satisfaction and create a more engaging experience.

% memory system evaluation
The quality check for memory system evaluation confirmed the effectiveness of our structured summarization approach, enabling the system to construct an accurate and structured Knowledge Graph (KG).
%% COMMENTS ON GRAPH VERSUS RAG GO HERE!!!!
In choosing between Knowledge Graphs (KGs) and Retrieval-Augmented Generation (RAG) for memory representation, our study found that both approaches exhibited comparable performance within our dataset size. A notable aspect is that even for specific queries, the graph-based approach achieves performance comparable to traditional RAG, despite not relying on extended summaries. Instead, it leverages a condensed representation—the graph itself—which effectively captures and organizes relevant information.
Moreover, the structured and explainable nature of KGs (and the possibility to explore them interactively through a dedicated GUI) provided distinct advantages in terms of fact verification, error diagnosis, and interpretability. Unlike RAG, which relies on probabilistic retrieval and may introduce opaque or untraceable errors, the explicit relational structure of a KG allows for easier debugging and controlled updates. 
%Why knowledge graphs???
As the size of interactions grow, the KG could facilitate shared learning across multiple users. By identifying patterns within training data—such as common physical challenges among users practicing the same sport—the robot could proactively recommend tailored exercises based on broader community insights.

%%% LIMITATIONS AND FUTURE WORK
%Scalability of Knowledge Representation: 
The scalability of our Knowledge Graph remains an open question. Further research is needed to determine the optimal number of examples required for efficient graph construction and to evaluate how retrieval performance scales with larger datasets. We relied on 28 interactions with simulated users for KG construction, but the ultimate goal is to deploy this system in longitudinal interactions. To prevent exponential KG growth, which could degrade retrieval performance, an optimal approach would be to implement dynamic graphs that intelligently manage memory updates and selective forgetting of obsolete or less relevant information. Developing such mechanisms will be a key focus of future work.

% \section{CONCLUSIONS}
% This study demonstrates that LLM-powered decision-making, combined with structured memory representation on different layers,
% %(procedural with tools, short term with WM prompt and long-term with graph)
% can significantly enhance social and task-oriented HRI. While our architecture successfully enables autonomous and adaptive interactions, further work is needed to investigate memory scalability, retrieval accuracy and adaptive interaction strategies. By addressing these challenges, we move closer to the goal of intelligent, socially aware robots capable of long-term, personalized engagement in human environments.

%Future research should focus on integrating temporal reasoning mechanisms to enable the robot to track progress over multiple encounters, anticipate user needs, and refine coaching strategies accordingly.

\addtolength{\textheight}{-12cm}   % This command serves to balance the column lengths

%%%%%%%%%%%%%%%%%%%%%%%%%%%%%%%%%%%%%%%%%%%%%%%%%%%%%%%%%%%%%%%%%%%%%%%%%%%%%%%%
\section*{ACKNOWLEDGMENT}
This work was supported by:
"RAISE – Robotics and AI for Socioeconomic Empowerment" and by the European Union - NextGenerationEU, National Recovery and Resilience Plan (NRRP). Luca Garello is member of the RAISE Innovation Ecosystem.
FAIR – Future Artificial Intelligence Research Foundation under the National Recovery and Resilience Plan (NRRP), MUR Call No. 341/2022, Extended Partnership on Artificial Intelligence – Foundational Aspects.
%Giulia Belgiovine is a member of the FAIR Ecosystem.
%%%%%%%%%%%%%%%%%%%%%%%%%%%%%%%%%%%%%%%%%%%%%%%%%%%%%%%%%%%%%%%%%%%%%%%%%%%%%%%%

%\begin{thebibliography}{99}
%\end{thebibliography}

 \bibliographystyle{IEEEtran} % use IEEEtran.bst style
\bibliography{IEEEabrv, IEEEexample}

\begin{thebibliography}{10}
\providecommand{\url}[1]{#1}
\csname url@rmstyle\endcsname
\providecommand{\newblock}{\relax}
\providecommand{\bibinfo}[2]{#2}
\providecommand\BIBentrySTDinterwordspacing{\spaceskip=0pt\relax}
\providecommand\BIBentryALTinterwordstretchfactor{4}
\providecommand\BIBentryALTinterwordspacing{\spaceskip=\fontdimen2\font plus
\BIBentryALTinterwordstretchfactor\fontdimen3\font minus \fontdimen4\font\relax}
\providecommand\BIBforeignlanguage[2]{{%
\expandafter\ifx\csname l@#1\endcsname\relax
\typeout{** WARNING: IEEEtran.bst: No hyphenation pattern has been}%
\typeout{** loaded for the language `#1'. Using the pattern for}%
\typeout{** the default language instead.}%
\else
\language=\csname l@#1\endcsname
\fi
#2}}

\bibitem{wang2024large}
J.~Wang, E.~Shi, H.~Hu, C.~Ma, Y.~Liu, X.~Wang, Y.~Yao, X.~Liu, B.~Ge, and S.~Zhang, ``Large language models for robotics: Opportunities, challenges, and perspectives,'' \emph{Journal of Automation and Intelligence}, 2024.

\bibitem{ZHANG2023100131}
\BIBentryALTinterwordspacing
C.~Zhang, J.~Chen, J.~Li, Y.~Peng, and Z.~Mao, ``Large language models for human–robot interaction: A review,'' \emph{Biomimetic Intelligence and Robotics}, vol.~3, no.~4, p. 100131, 2023. [Online]. Available: \url{https://www.sciencedirect.com/science/article/pii/S2667379723000451}
\BIBentrySTDinterwordspacing

\bibitem{incao2025roadmap}
S.~Incao, C.~Mazzola, G.~Belgiovine, and A.~Sciutti, ``A roadmap for embodied and social grounding in llms,'' in \emph{Social Robots with AI: Prospects, Risks, and Responsible Methods}.\hskip 1em plus 0.5em minus 0.4em\relax IOS Press, 2025, pp. 43--52.

\bibitem{irfan2025between}
B.~Irfan, S.~Kuoppam{\"a}ki, A.~Hosseini, and G.~Skantze, ``Between reality and delusion: challenges of applying large language models to companion robots for open-domain dialogues with older adults,'' \emph{Autonomous Robots}, vol.~49, no.~1, p.~9, 2025.

\bibitem{hatalis2023memory}
K.~Hatalis, D.~Christou, J.~Myers, S.~Jones, K.~Lambert, A.~Amos-Binks, Z.~Dannenhauer, and D.~Dannenhauer, ``Memory matters: The need to improve long-term memory in llm-agents,'' in \emph{Proceedings of the AAAI Symposium Series}, vol.~2, no.~1, 2023, pp. 277--280.

\bibitem{zhang2024survey}
Z.~Zhang, X.~Bo, C.~Ma, R.~Li, X.~Chen, Q.~Dai, J.~Zhu, Z.~Dong, and J.-R. Wen, ``A survey on the memory mechanism of large language model based agents,'' \emph{arXiv preprint arXiv:2404.13501}, 2024.

\bibitem{pan2024unifying}
S.~Pan, L.~Luo, Y.~Wang, C.~Chen, J.~Wang, and X.~Wu, ``Unifying large language models and knowledge graphs: A roadmap,'' \emph{IEEE Transactions on Knowledge and Data Engineering}, vol.~36, no.~7, pp. 3580--3599, 2024.

\bibitem{10.1145/3610977.3634966}
\BIBentryALTinterwordspacing
C.~Y. Kim, C.~P. Lee, and B.~Mutlu, ``Understanding large-language model (llm)-powered human-robot interaction,'' in \emph{Proceedings of the 2024 ACM/IEEE International Conference on Human-Robot Interaction}, ser. HRI '24.\hskip 1em plus 0.5em minus 0.4em\relax New York, NY, USA: Association for Computing Machinery, 2024, p. 371–380. [Online]. Available: \url{https://doi.org/10.1145/3610977.3634966}
\BIBentrySTDinterwordspacing

\bibitem{clabaugh2019escaping}
C.~Clabaugh and M.~Matari{\'c}, ``Escaping oz: Autonomy in socially assistive robotics,'' \emph{Annual Review of Control, Robotics, and Autonomous Systems}, vol.~2, no.~1, pp. 33--61, 2019.

\bibitem{10.1145/3712265}
\BIBentryALTinterwordspacing
M.~Spitale, M.~Axelsson, and H.~Gunes, ``Vita: A multi-modal llm-based system for longitudinal, autonomous, and adaptive robotic mental well-being coaching,'' \emph{J. Hum.-Robot Interact.}, Jan. 2025, just Accepted. [Online]. Available: \url{https://doi.org/10.1145/3712265}
\BIBentrySTDinterwordspacing

\bibitem{kang2024nadine}
H.~Kang, M.~Ben~Moussa, and N.~M. Thalmann, ``Nadine: A large language model-driven intelligent social robot with affective capabilities and human-like memory,'' \emph{Computer Animation and Virtual Worlds}, vol.~35, no.~4, p. e2290, 2024.

\bibitem{ali2024robots}
H.~Ali, P.~Allgeuer, C.~Mazzola, G.~Belgiovine, B.~C. Kaplan, L.~Gajdo{\v{s}}ech, and S.~Wermter, ``Robots can multitask too: Integrating a memory architecture and llms for enhanced cross-task robot action generation,'' in \emph{2024 IEEE-RAS 23rd International Conference on Humanoid Robots (Humanoids)}.\hskip 1em plus 0.5em minus 0.4em\relax IEEE, 2024, pp. 811--818.

\bibitem{maharana2024evaluatinglongtermconversationalmemory}
\BIBentryALTinterwordspacing
A.~Maharana, D.-H. Lee, S.~Tulyakov, M.~Bansal, F.~Barbieri, and Y.~Fang, ``Evaluating very long-term conversational memory of llm agents,'' 2024. [Online]. Available: \url{https://arxiv.org/abs/2402.17753}
\BIBentrySTDinterwordspacing

\bibitem{belgiovine2022towards}
G.~Belgiovine, J.~Gonzalez-Billandon, G.~Sandini, F.~Rea, and A.~Sciutti, ``Towards an hri tutoring framework for long-term personalization and real-time adaptation,'' in \emph{Adjunct Proceedings of the 30th ACM Conference on User Modeling, Adaptation and Personalization}, 2022, pp. 139--145.

\bibitem{metta2006yarp}
G.~Metta, P.~Fitzpatrick, and L.~Natale, ``Yarp: yet another robot platform,'' \emph{International Journal of Advanced Robotic Systems}, vol.~3, no.~1, p.~8, 2006.

\bibitem{mccrae1992introduction}
R.~R. McCrae and O.~P. John, ``An introduction to the five-factor model and its applications,'' \emph{Journal of personality}, vol.~60, no.~2, pp. 175--215, 1992.

\bibitem{gao2023retrieval}
Y.~Gao, Y.~Xiong, X.~Gao, K.~Jia, J.~Pan, Y.~Bi, Y.~Dai, J.~Sun, H.~Wang, and H.~Wang, ``Retrieval-augmented generation for large language models: A survey,'' \emph{arXiv preprint arXiv:2312.10997}, vol.~2, 2023.

\end{thebibliography}


% \subsection{List of Tools Accessible by the Interaction Manager}
% \label{appendix:list_of_tools}

% Each tool consists of a title (ideally self-explanatory) and a brief description of its intended use. Tools can be customized for specific robotic tasks and developed based on domain-specific knowledge. The tools designed for our particular use case include:
% \begin{itemize}
%     \item \texttt{chatting}
%     \item \texttt{user\_profiling}
%     \item \texttt{plan\_yoga\_session}
%     \item \texttt{show\_the\_pose}
%     \item \texttt{ask\_final\_feedback}
%     \item \texttt{user\_name\_retriever}
%     \item \texttt{goodbye}
% \end{itemize}

%%%%%%%%%%%%%%%%%%%%%%%%%%%%%%%%%%%%%%%%%%%%%%%%%%%%%%%%%%%%%%%%%%%%%%%%%%%%%%%%

% \subsection{Evaluation of the HRI architecture during the user study}

%%%%%%%%%%%%%%%%%%%%%%%%%%%%%%%%%%%%%%%%%%%%%%%%%%%%%%%%%%%%%%%%%%%%%%%%%%%%%%%%

% \subsection{Example of a Bio for fake users}
% \label{appendix:bio_fake_users}

% \begin{tcolorbox}[colframe=gray!70, colback=gray!10, boxrule=0.4pt, coltitle=black]
% \textbf{Name}: Alex Novak  \\
% \textbf{Age}: 32 \\
% \textbf{Interests}: Cycling, photography, and science fiction \\
% \textbf{Sport}: Enjoys cycling on weekends. \\
% \textbf{Job}: Teacher\\
% \textbf{Personality}:\\
% - Openness: High \\
% - Conscientiousness: Moderate \\
% - Extraversion: Moderate \\
% - Agreeableness: Moderate \\
% - Neuroticism: Moderate \\
% \textbf{Probability of yoga success}: 85\% 
% \end{tcolorbox}

\end{document}